\documentclass[letterpaper, 10 pt, conference]{ieeeconf}  

\IEEEoverridecommandlockouts                              

\usepackage{graphics} 
\usepackage{epsfig} 
\usepackage{times} 
\usepackage{amsmath} 
\usepackage{amssymb}  
\usepackage{multirow}
\usepackage{booktabs} 
\usepackage{adjustbox}
\usepackage{spverbatim}
\usepackage{caption}
\usepackage{subcaption}
\usepackage[noadjust]{cite}
\usepackage{balance}

\usepackage{tikz}
\usepackage{booktabs}
\usepackage{comment}

\newcommand{\mypara}[1]{\vspace{0.5em}\noindent\textbf{#1}}

\newcommand{\LTLf}{{\textsf{LTL}$_f$}}

\title{\LARGE \bf
Embedding Symbolic Temporal Knowledge into Deep Sequential Models
}
\author{Yaqi Xie$^*$, Fan Zhou$^*$, and Harold Soh\\Dept. of Computer Science, National University of Singapore.\\{\small\texttt{\{yaqixie, zhoufan ,harold\}@comp.nus.edu.sg}}
\thanks{$^*$Equal Contribution.}
}

\begin{document}

\maketitle
\thispagestyle{empty}
\pagestyle{empty}

\begin{abstract}
\label{sec:abstract}
Sequences and time-series often arise in robot tasks, e.g., in activity recognition and imitation learning. In recent years, deep neural networks (DNNs) have emerged as an effective data-driven methodology for processing sequences given sufficient training data and compute resources. However, when data is limited, simpler models such as logic/rule-based methods work surprisingly well, especially when relevant prior knowledge is applied in their construction. However, unlike DNNs, these  ``structured'' models can be difficult to extend, and do not work well with raw unstructured data. In this work, we seek to learn flexible DNNs, yet leverage prior temporal knowledge when available. Our approach is to embed symbolic knowledge expressed as linear temporal logic (LTL) and use these embeddings to guide the training of deep models. Specifically, we construct semantic-based embeddings of automata generated from LTL formula via a Graph Neural Network. Experiments show that these learnt embeddings can lead to improvements on downstream robot tasks such as sequential action recognition and imitation learning. 
\end{abstract}
\section{Introduction}
\label{sec:introduction}

Sequence learning is crucial building-block for AI and robotics; it is applied to various problems  including action prediction and policy learning. 
Significant advances have been made in sequence learning, exemplified by improvements on tasks ranging from visual tracking~\cite{wu2015} and language translation~\cite{Vaswani2017} to human modeling~\cite{soh2019multi} and game playing~\cite{Silver2017}. This progress has been largely powered by deep learning~\cite{LeCunBH15}. However, deep models often require large amounts of training data, which may limit their application in situations where data is not as readily available. For example, in imitation learning contexts, expert demonstrations are typically expensive and difficult to procure. 


In many settings, high-level structured knowledge is often available in addition to data. Consider that humans teach children not only by giving examples, but also through structured information. For example, a parent who is assembling a Lego toy with child may point out that ``\emph{two blocks can be put together to form a bigger block},'' or ``\emph{according to the picture, the red brick should be on the left}''. Similarly, cooking recipes include general tips such as the following when making apple pie: ``\emph{to cook apples thoroughly, first bring the water to a simmer, and after lowering in the apples, cover the pot with a baking parchment}''. However, it remains unclear how we can best leverage these types of structured knowledge in deep neural networks (DNNs). 

In this work, we seek to incorporate \textit{temporal knowledge} into deep sequential learning. We focus on temporal knowledge expressed in finite Linear Temporal Logic (\LTLf{}). \LTLf{} is well-defined and unambiguous compared to natural language, yet relatively easy for humans to derive and interpret. These properties make \LTLf{} a useful language for specifying temporal relationships and dynamic constraints, and for automated reasoning. As such, various attempts have been proposed to incorporate \LTLf{} in sequential models, e.g., in reinforcement learning via planning~\cite{fainekos2005,kress2007, karaman2011} and by using logic checkers as auxiliary losses~\cite{Wen2017, Hasanbeig2018, Littman2017}. 


Different from prior work, we explore the incorporation of \LTLf{} knowledge into neural networks via  \emph{real-vector embeddings}. 
In contrast to symbols and their operators, embeddings are naturally processed by standard deep neural networks. When properly structured/learned, distances (e.g., Euclidean) in the embedding space can be exploited to quickly compute whether a given output from a deep network is consistent with related \LTLf{} formulae. Unlike standard model checkers, our approach enables a form of ``soft''-satisfiability, which can potentially generalize knowledge. In our cooking example, when making a fruit pie, the knowledge for using apples (e.g., cutting, poaching) may  apply to other fruits (e.g, pears). 
However, these advantages are contingent upon a successful transfer of the information contained within \LTLf{} formulae into corresponding real-vectors. A principal challenge is that \LTLf{} syntax trees are multifarious; the same formulae can often be written in many different ways, which makes it hard for neural networks to learn relevant semantics. 


\begin{figure*}
      \centering
      \includegraphics[width = 0.95\textwidth]{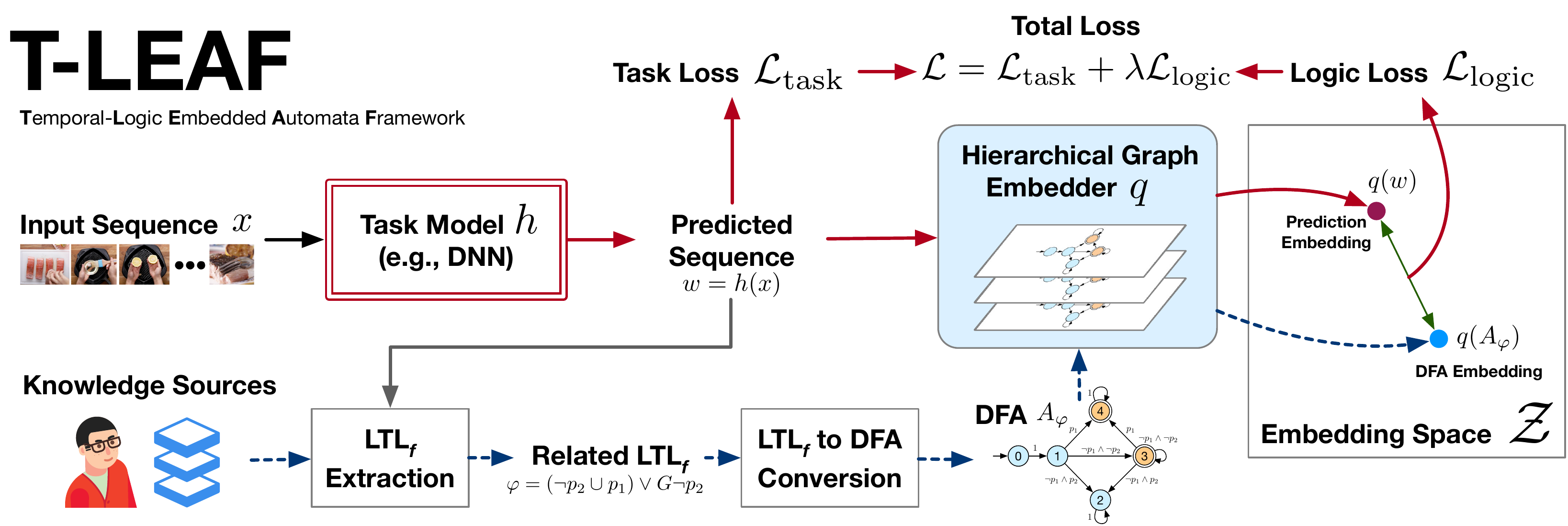}
      \caption{\small An overview of the \textbf{T}emporal-\textbf{L}ogic \textbf{E}mbedded \textbf{A}utomata \textbf{F}ramework (\textbf{T-LEAF}). T-LEAF trains a deep task model $h$ (red box) to be consistent with prior knowledge encoded as \LTLf{} formulae. The extracted \LTLf{} formulae are first converted to a DFA $A_\varphi$. Both the DFA $A_\varphi$ and the predicted sequence $w$ (from the task model $h$) are projected to their respective embeddings using a hierarchical graph embedder $q$. The two embeddings, $z_\varphi = q(A_\varphi)$ and $z_w = q(w)$, are compared to give a logic loss $\mathcal{L}_\textrm{logic}$, which is back-propagated to the task model together with the task loss $\mathcal{L}_\textrm{task}$ (by following the solid red arrows in reverse). In effect, $\mathcal{L}_\textrm{logic}$ provides an additional training signal that encourages the model $h$ to produce outputs compatible with existing knowledge.}
      \label{fig:T-LEAF}
\end{figure*}

Our insight is that a \emph{deterministic finite automaton} (DFA) (translated from an \LTLf{} specification) incorporates the needed semantics, yet has a structure that is amenable to learning via graph neural networks (GNN)~\cite{Wu2020,Zhou2018}. In particular, we note that the message propagation between nodes in graph convolutional layers can approximate the path traversals used in \LTLf{} model checking algorithms. 
Indeed, our model-checking experiments in Sec.~\ref{sec:experiment_synthetic} show that embedding DFAs resulted in far better performance compared to syntax trees. 

Based on this idea, we take first steps towards a practical approach for training deep sequential models using a combination of data and prior temporal knowledge. 
We propose the Temporal-Logic Embedded Automata Framework (T-LEAF) that ``steers'' deep models during training to be consistent with specified \LTLf formulae (Fig. \ref{fig:T-LEAF}). Specifically, T-LEAF uses a custom-designed hierarchical embedder to convert DFAs into embeddings that are then used in a \emph{logic loss}. The logic loss encourages the output of models to be consistent with the embedding, and hence the \LTLf{} specification. Our experiments on two tasks in the robot cooking domain give positive evidence that T-LEAF is able to improve deep models. Specifically, we tested T-LEAF on sequential human activity recognition and imitation learning. In both tasks, training with T-LEAF led to better performance for baseline and state-of-the-art deep networks. 

In summary, the main contribution of this paper is a new framework for utilizing symbolic temporal logic in sequential deep models. Experiments show T-LEAF is able to enhance deep models on two challenging tasks. To our knowledge, this is the first work to study graph-based embeddings of \LTLf{} formulae, and to demonstrate their usefulness for training better models. We believe further improvements are possible by fine-tuning T-LEAF. More broadly, our results indicate that temporal logic embeddings are promising and warrant further research.

\section{Preliminaries: Linear Temporal Logic}
\label{sec:related_works}

Linear Temporal Logic (\textsf{LTL}) is a propositional modal logic often used to express temporally extended constraints over state trajectories~\cite{Pnueli1977}. 
In this work, we use \textsf{LTL} interpreted over finite traces, which is also called \textit{finite} \textsf{LTL} (\LTLf).
\LTLf{} has a natural and intuitive syntax. As a formal language, it has well-defined semantics and thus is unambiguously interpretable, which is an advantage over using natural language directly as auxiliary information~\cite{Goyal2019}\footnote{It is also possible to translate natural language into \LTLf{}~\cite{Dzifcak2009}, which can then be analyzed using developed tools (e.g., model checkers) and embedded using T-LEAF.}. In this section, we briefly review relevant background on \LTLf{}; we focus on material relevant to our method and refer readers to excellent review articles~\cite{Baier2008, Rozier2011,French2013} for more  information.

\mypara{Finite Linear Temporal Logic.} The syntax of \LTLf{} for a finite set of propositions $p \in AP$ includes the standard logic connectives ($\land, \lor, \neg$), \textsf{true} and \textsf{false} symbols, and temporal operators \textit{next} (\textsf{X}) and \textit{until} (\textsf{U}), 
\begin{align*}
    \varphi := \textsf{false} | \textsf{true} | p | \neg p| \varphi_1 \land \varphi_2 | \varphi_1 \lor \varphi_2| \neg \varphi| \textsf{X} \varphi| \varphi_1 \textsf{U} \varphi_2
\end{align*}
Finite \LTLf{} formulae are interpreted over finite traces $w=\sigma_0\sigma_1...\sigma_n$ of propositional states, where each $\sigma_i$ is a set of propositions from ${P}$ that are true in $\sigma_i$. We say that $w$ \textit{satisfies} a formula $\varphi$, denoted $w\models\varphi$, when $w,0\models\varphi$, where:
\begin{align}
     w, i \models & p \text{ iff } p\in\text{ AP and } \sigma_1 \models p\\
     w, i \models &\neg \alpha \text{ iff } w, i \not\models \alpha\\
      w, i \models &\alpha \land \beta \text{ iff } w, i \models \alpha \text{ and } w, i \models \beta\\
      w, i \models& \textsf{ X } \alpha \text{ iff } w, i+1 \models \alpha\\
     w, i \models& \alpha \textsf{ U } \beta \text{ iff } w, k \models \beta \text{ for some } i \leq k < n \\ &\text{ and } w, j \models \alpha \text{ for all } i\leq j < k
\end{align}

\LTLf{} has been used in AI and robot planning to specify temporally-extended goals~\cite{Bacchus2000, Baier2006, Camacho2017} and preferences~\cite{Baier2009, Bienvenu2011}. 
Compared to low-level robot programming, specifying (and interpreting) high-level task requirements using  \LTLf{} is relatively easy for humans.  To define a task, \LTLf{} needs a high-level domain specific vocabulary comprising a set of propositions that relates to properties of the environment, or the occurrence of events that can be determined to be \textsf{true} or \textsf{false}. 

\mypara{LTL and Automata. }
For every \LTLf{} formula $\varphi$, we can construct a deterministic finite-state automaton (DFA), $A_\varphi$, which accepts the models of $\varphi$~\cite{Zhu2017, Alberto2018}, i.e. the interpretations that satisfy $\varphi$. 
A DFA is a tuple $A_\varphi = \langle O, \Sigma, o_0, \delta, \alpha\rangle$, where $O$ is a finite set of states, $\Sigma$ contains all subsets of propositions $\varphi, o_0 \in O$ is the initial state, $\delta\subseteq O \times L(P) \times O$ is a transition relation, where $L(P)$ is the set of propositional formulae over $P$, and $\alpha\subseteq O$ is a set of accepting states. A run of $A_\varphi$ on a word $w=x_1...x_n\in\Sigma^*$ is a sequence of states $o_0o_1...o_n$ such that $(o_{i-1}, \varphi_i, o_i) \in \delta$ and $x_i\models\varphi_i$ for each $i\in{1,...n}$. A run is accepting if $o_n\in\alpha$. 

A DFA $A_\varphi$ can be represented as a directed graph $\mathcal{G}_\varphi=(\mathcal{V,E})$ with nodes $v_i\in\mathcal{V}$ representing the automata states, and directed edges $e_j = (v_s, v_t, \mathsf{F}_{s,t}) \in\mathcal{E}$ from a source node $v_s\in \mathcal{V}$ to a target node $v_t\in \mathcal{V}$. Each edge $e_j$ contains a propositional logic formula $\mathsf{F}_{s,t}$ (also denoted $\mathsf{F}_{j}$) that defines the conditions under which transit is permitted from $v_s$ to $v_d$. There are three types of node states: an initial state, intermediate states, and acceptance states 
A given trace is satisfying trace if it begins at the initial state and terminates at one of the acceptance states. For the remainder of this paper, we will assume DFAs are represented as directed graphs. 

\section{Method: Embedding DFAs for Deep Model Training via T-LEAF}
\label{sec:method}

This section details our primary contribution, i.e., a framework for using symbolic temporal knowledge in the training of deep models. An overview of our proposed Temporal-Logic Embedded Automata Framework (T-LEAF) is shown in Fig.  \ref{fig:T-LEAF}. Briefly, the key component in our framework is a hierarchical graph embedder $q$ that embeds DFAs and predicted sequences from deep models (traces) into a shared real-vector space $\mathcal{Z} \subseteq \mathbb{R}^d$. We train $q$ such that   
embedded formulae are closer to their satisfying traces. We can then easily evaluate a \emph{logic loss} --- that captures show much a given trace satisfies related formulae (and hence the symbolic knowledge) --- by computing distances in $\mathcal{Z}$. In the following, we first detail the DFA embedder, followed by the logic loss.


\mypara{Hierarchical DFA Embedder. }
Recall that a DFA can be represented as a directed graph $\mathcal{G}_\varphi=(\mathcal{V,E})$ with three node types and propositions along the edges. To embed the information contained in the DFA, we need to embed the overall graph structure, together with the propositions. 

We propose a \emph{hierarchical embedder} $q$ that comprises an edge-embedder $q_e$ and a meta-embedder $q_m$. Embedding a given DFA involves the construction of an intermediate graph $\widehat{\mathcal{G}}_\varphi$ that will be embedded using $q_m$. At a high-level, our goal is to convert each edge formula into a corresponding \emph{node feature vector}\footnote{GNNs can process edge features but we found using node features led to better empirical performance.}  The key steps are:
\begin{enumerate}
    \item Embed each proposition $\mathsf{F}_{j}$ along edge $e_j$ into a corresponding vector $s_j = q_e(\mathsf{F}_{j})$. 
    \item Construct $\widehat{\mathcal{G}}_\varphi = (\widehat{\mathcal{V}}, \widehat{\mathcal{E}})$, where $\widehat{\mathcal{V}}$ contains all nodes in the original graph $\mathcal{G}$. 
    \item For each edge $e_j = (v_s, v_d) \in \mathcal{E}$ (in the original DFA graph), create a new node $\widehat{v}_{e_j}$ in $\widehat{\mathcal{V}}$, and add edges $(v_s, \widehat{v}_{e_j})$ and $(\widehat{v}_{e_j}, v_d)$ to $\widehat{\mathcal{E}}$. 
    \item Embed the graph $\widehat{\mathcal{G}}$ using the meta-embedder $q_m$
    \item Obtain the entire graph representation by aggregating node embeddings sampled via random walks initiated from the initial state node~\cite{Goyal2018}.
\end{enumerate}
In this work, both edge-embedder $q_e$ and meta-embedder $q$ are multi-layer Graph Convolutional Networks (GCNs)~\cite{Kipf2017} with four layers.

\mypara{Embedder Training. } To train our meta embedder, we use a \emph{triplet loss} that encourages formulae embeddings to be close to satisfying traces, and far from unsatisfying traces. Let $z_\varphi = q(A_\varphi)$ be the DFA embedding. Define $ z_\mathsf{T} = q(w_\mathsf{T})$ and $z_\mathsf{F} = q(w_\mathsf{F})$ as the trace embeddings for a satisfying and unsatisfying trace, respectively. Note that the graph structures for traces are linear; edges are a conjunction of propositions at each respective time-step. Our triplet loss is a hinge loss:
\begin{equation}
\label{eq-marginLoss}
\begin{split}
\ell_\textrm{triplet}(A_\varphi, w_\mathsf{T}, w_\mathsf{F}) =
\max \{d(z_\varphi, z_\mathsf{F}) - d(z_\varphi, z_\mathsf{T}) + m, 0\}, 
\end{split}
\end{equation}
where $ d(x,y) $ is the squared Euclidean distance between $ x $ and $ y $, and $ m $ is the margin. Training the embedder entails optimizing a combined loss:
\begin{equation}
	L_{\textrm{emb}} = \sum_{A_\varphi}\sum_{ w_\mathsf{T}, w_\mathsf{F}} \ell_\textrm{triplet}(A_\varphi, w_\mathsf{T}, w_\mathsf{F}),
	\label{eq-loss}
\end{equation}
where the summation is over formulas and associated pairs of satisfying and unsatisfying traces in our dataset. In practice, pairs of traces are randomly sampled for each formula during training. In our experiments, the edge-embedder $q_e$ is trained first (and fixed), followed by the meta-embedder $q_m$. We found this approach to improve training stability, and leave alternative training schemes (e.g., joint or interleaved) to future work.  



\mypara{The Logic Loss. }
We train a given target task model $h$ by augmenting its per-datum task loss with a logic loss $\ell_{\textrm{logic}}$, 
\begin{align}
\ell = \ell_{\textrm{task}} + \lambda \ell_{\textrm{logic}},
\end{align}
where 
$\ell_{\textrm{logic}} = \|q(A_\varphi) - q( h(x) )\|_2^2 $ is the embedding distance between the DFA (converted from \LTLf{} formula related to the input $x$) and the predicted output sequence $w = h(x)$. The task-specific loss 
$\ell_{\textrm{task}}$ depends on the application, e.g., cross-entropy for classification. Lastly, $ \lambda $ is hyper-parameter that trades-off task performance and compliance with prior knowledge. 
\section{Related Work}
\label{sec:related_work}

T-LEAF is related to a body of work on embedding symbolic logic for prediction~\cite{xie2019, Xu2018, Kim2016, allamanis2016, Tai2015, Le2015, Zhu2015} and reasoning tasks~\cite{Lee2020, Wang2017, Paliwal2019}. The key difference is that past work has largely focused on general  propositional and first-order logic. In contrast, T-LEAF embeds temporal logic, which is useful for many robotics applications.

Prior work using \LTLf for robotics has focused primarily on planning~\cite{fainekos2005, kress2007, karaman2011, fainekos2009}---e.g., synthesizing controllers from the product of \LTLf{} automata and environment automata---and using temporal-logic checkers to shape reward functions~\cite{Wen2017,Wen2018,Hasanbeig2018,Li2017, Littman2017, Icarte2018}. Recently, a differentiable \textsf{LTL} loss was proposed in \cite{Innes2020}. However, it is limited to continuous control or regression tasks; in binary symbolic domains, the loss is non-differentiable and equivalent to a model checker. 

Different from the work above, T-LEAF is designed to improve ``target'' deep sequential models by embedding prior knowledge. Compared to a direct application of a model checker, we posit that learned embeddings may yield a more informative loss and gradients; intuitively, the gradients may provide ``directional'' information towards regions of complying models. To our knowledge, the T-LEAF logic loss is only differentiable loss for (binary) propositional temporal logic.




\section{Experiment: Model Checking}
\label{sec:experiment_synthetic}
In this section, we focus on testing whether deterministic finite-state automata (DFA) are more amenable to embedding compared to \LTLf{} syntax trees. Specifically, we conduct an experiment using a model checking problem: given the embedding of a \LTLf{} formula $\varphi$ and the embedding of a trace $w$, predict whether $w$ models $\varphi$. 

\mypara{Dataset and Experiment Setup. }
The complexity of formulae is (coarsely) reflected by its number of propositions $n_v$ and the syntax tree width $w_t$. As such, we synthesized three datasets with $(n_v, w_t) = (3,10), (3,20), (6,20)$, corresponding to ``Low'', ``Moderate'' and ``High'' complexities, respectively. Formulae were translated into DFA using LTLfKit~\cite{Zhu2017,Alberto2018}. 

Our T-LEAF hierarchical embedder uses 4 graph convolution layers, with 100 hidden units per layer, and outputs a 200-dimension embedding for each input automata or trace. The neural network used for classification is a multi-layer perceptron with 512 and 128 hidden units in the first and second layers, respectively.

\begin{figure}
    \centering
    \includegraphics[width=1.0\columnwidth]{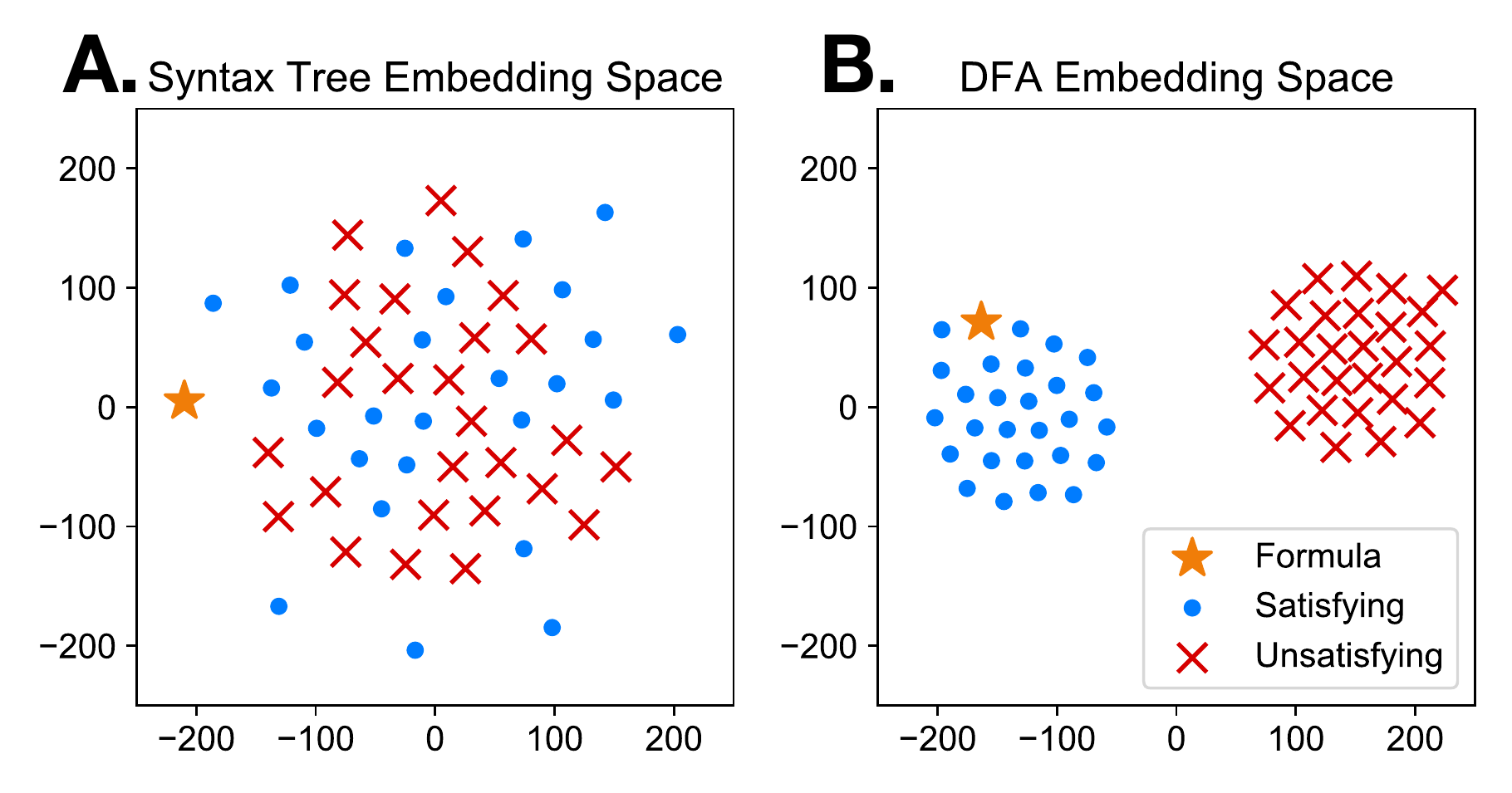}
    \caption{\small t-SNE 2D-projection of the learnt embedding spaces for (\textbf{A.}) Syntax Tree inputs, and (\textbf{B.}) DFA inputs. The DFA space shows a clear separation between satisfying traces (blue dots) and unsatisfying traces (red crosses) for the shown formula (star).}
    \label{fig:embedding_space}
    \vspace{-0.3cm}
\end{figure}

\mypara{Results. }
Prediction accuracies with standard errors over three independent runs are reported in Table \ref{tab:synthetic_result}. Across the datasets, we observe that using DFA graphs results in far better accuracy compared to \LTLf{} syntax trees; the difference in performance is large $\approx 16\%-27\%$. This difference can be explained by examining the embedding spaces;  Fig.~\ref{fig:embedding_space} illustrates a 2D projection of a sample formula and its satisfying/unsatisfying traces. The Syntax Tree embeddings appear to lack structure. In contrast, there is a clear separation between satisfying traces and unsatisfying traces in the DFA embedding space, and the satisfying traces are closer to the formula. These results support our hypothesis that DFA can more easily embedded using graph neural networks, compared to syntax trees.




\begin{table}
\centering
\caption{\small Prediction Accuracy (with Std. Error) on Synthetic Datasets with Varying Complexity. Highest Accuracies in \textbf{Bold}.}
    \begin{tabular}{c|c c c}
    \hline
    \hline 
         \multirow{2}{*}{Embedder Input}  & \multicolumn{3}{c}{Accuracy} \\ 
		 &  Low & Moderate & High \\ \hline 
         Syntax tree&  56.16 (0.92)& 54.13 (0.62) & 50.00 (0.00)\\ 
        \hline 
         DFA & \textbf{83.43} (0.73) &\textbf{78.43} (1.54) & \textbf{66.90} (0.70)\\ 
    \hline 
    \hline 
    \end{tabular}
    \label{tab:synthetic_result}
      \vspace{-0.2cm}
\end{table}

\section{Sequential Human Action Recognition}
\label{sec:experiment_video}
In this section, we show how T-LEAF can be applied to Sequential Human Action Recognition. This task is important in many contexts, e.g., a robot needs recognize human actions to assist appropriately. We focus on a cooking domain where a model is trained to predict an (\textit{action}, \textit{object}) pair sequentially given temporal visual information (i.e., image frames). 
Target models could benefit from temporal common-sense knowledge, such as the affordances of certain objects and the likely ordering between actions. Our goal was to establish if incorporating such prior knowledge via T-LEAF results in better task models. 

\mypara{Dataset. }
We evaluated our method on the \textit{Tasty Video Dataset}~\cite{Sener2019}, which contains 4027 recipe videos involving 1199 unique ingredients.
Each recipe is self-contained with an ingredient list, step-wise instructions with temporal alignment, and a video demonstrating the preparation. The Tasty Videos are captured with a fixed overhead camera and focus entirely on preparation of the dish. 
However, target labels ((\textit{action}, \textit{ingredient}) pairs for each step) were not provided, so we crowd-sourced labels for 500 videos. To balance the labels, we trimmed  the number of action and ingredient classes (to 64 actions and 155 ingredients) by removing low frequency classes and combining similar classes, such as \textit{cheddar} and \textit{cheese}.

\mypara{Temporal Knowledge. }
We used {(\textit{action}, \textit{ingredient})} pairs as propositions, such as {(\textit{grill}, \textit{salmon})}, which is \textsf{true} at time step $t$ if it exists in the $t^{\textrm{th}}$ step. Two type of rules are defined over the  propositions:
\begin{itemize}
    \item \textit{Affordance Constraints.} These constraints helps to eliminate action and ingredient pairs where the action is not afforded by the object, such as {(\textit{cut}, \textit{milk})}.
    \item \textit{Ordering Constraints.} Certain actions must come before other actions if they appear in the same video and are applied on the same object. For example, {(\textit{rinse}, \textit{cabbage})} should take place before {(\textit{cook}, \textit{cabbage})}. Note that there is no constraint if only one of pairs takes place, or they are not applied on the same object.
\end{itemize}
We extract affordance constraints and ordering constraints from the complete dataset to form our knowledge base \textsf{K}. 

\begin{figure*}
      \centering
      \includegraphics[width = .98\textwidth]{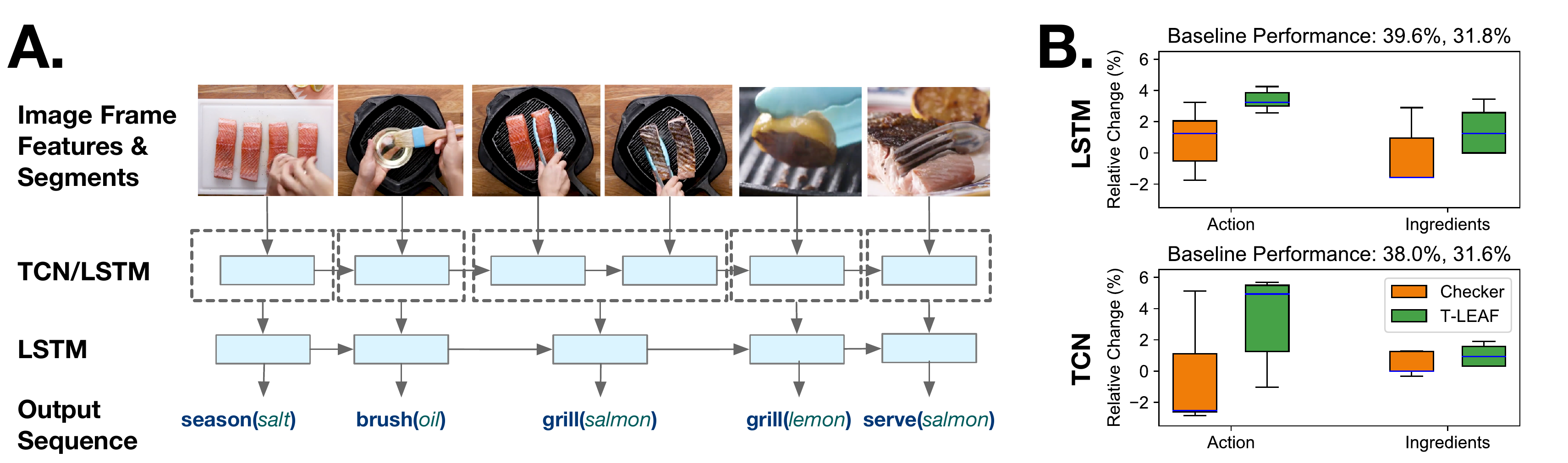}
      \caption{\small Sequential Human Action Recognition Experiment. (\textbf{A.}) An overview of the target model. We use ResNet-50~\cite{he2016} to extract visual features of each video frame, which are aggregated using a LSTM/TCN. The model is topped with two independent bi-directional LSTMs that predict the action and ingredient pair. (\textbf{B.}) Boxplots of relative changes after applying T-LEAF to reference models. T-LEAF improves performance on both the LSTM and TCN target models, as indicated by the positive changes, and generally outperforms target model training using a model checker.}
      \label{fig:tasty}
\end{figure*}


\mypara{Hierarchical Embedder. } The meta embedder $q$ used was a four-layer GCN with the same structure described in Sec.~\ref{sec:experiment_synthetic}. The edge embedder $q_e$ was two-layer GCN (256 hidden units per layer) that outputs a 100-dimensional embedding. 

We trained the hierarchical embedder using \LTLf{} formulae constructed from \textsf{K}. 
For each sample $x_i$, we constructed a \LTLf{} formula $\varphi_i$ that only contained propositions and constraints related to current video sample. Specifically, we enumerated all {(\textit{action}, \textit{ingredient})} pairs in current video, and extracted relevant constraints from \textsf{K} if all propositions, i.e., action ingredient pairs, involved in the constraint appeared in the video. The constraints were then combined together to form an \LTLf{} formula $\varphi_i$ specific to the video. $\varphi_i$ was then translated into a DFA $A_{\varphi_i}$. The edges of $A_{\varphi_i}$ are propositional formulae in Conjunctive Normal Form (CNF), which were fed into edge embedder $q_e$. To obtain the vector representation of a given proposition, we concatenated the GLoVe~\cite{Pennington2014Glove}
embeddings of the action and ingredient words (100-dimensions each), resulting an 200-dimension vector for each proposition. The edge embedder $q_e$ was trained first using the triplet loss described in Sec.~\ref{sec:method}, and then fixed to generate edge features of DFA graph $A_{\varphi_i}$ during the training of meta embedder $q$.

\mypara{Target Model. } 
In this experiment, the target model $h$ is a sequential human action recognition model (Fig.~\ref{fig:tasty}.A.). As visual segmentation was not the main focus of our work, we assumed the temporal alignment boundaries as given. We tested two slightly different models that use either a bi-directional LSTM ~\cite{hochreiter1997} (with 1024 hidden units) or a temporal convolution network (TCN)~\cite{bai2018} to aggregate the video features. The TCN is a multi-layer one-dimension convolutional neural network (conv1d) with different dilation factors across layers; it serves as a representative of a state-of-the-art model for temporal data. The TCN in our experiment has 3 channels of conv1d with 800 hidden units and kernel size 2. 

The input to the model $h$ is the video segment feature sequence. Suppose the $j^{\textrm{th}}$  video segment $c_j$ is composed of $L$ frames, i.e. $c_j = \{f^j_t\}_{t=1,2,...,L}$. Each frame $f^j_t$ is represented as a feature vector (2048-dimensions), which is the output of last fully-connected layer before the softmax layer in ResNet-50~\cite{he2016}. We obtain the segment vector $z_j$ by applying max pooling on output of LSTM or TCN. 
Then, $z_j$ is passed to two independent one-layer bidirectional LSTMs for action and ingredient classification. 
The prediction generated by model at each time step $j$ is an action-ingredient pair $(a_j, o_j)$. The complete prediction sequence for a video sample $i$ is $w_i = \{(a_1^i, o_1^i), (a_2^i, o_2^i), ..., (a_n^i, o_n^i)\}$, where $n$ is the number of steps in the video.

\mypara{Target Model Training via T-LEAF. } Training the target model $h$ using T-LEAF is  straightforward given the hierarchical embedder $q$. For each input $i$, we simply compute the logic loss $$\ell_{\textrm{logic}} = \|q(A_{\varphi_{w_i})}-q(w_i)\|^2_2.$$ The  DFA $A_{\varphi_{w_i}}$ was obtained by extracting the constraints related to the predicted sequence $w_i$ from the knowledge base \textsf{K} (in a similar manner as the training data for the hierarchical embedder). We then optimized the total loss $\ell = \ell_{\textrm{task}} + \lambda \ell_{\textrm{logic}}$, where the $\ell_{\textrm{task}}$ is the cross entropy loss and we set $\lambda = 5.0$. Optimization was carried out using Adam~\cite{kingma2014} with learning rate $10^{-4}$.

\mypara{Results. }
We compared the relative change in accuracy between five pairs of models; each pair was initialized with the same random seed, but trained using only the task loss (the reference model) or using the task loss together with the T-LEAF logic loss or a model-checker loss. Fig. ~\ref{fig:tasty}.B. summarizes our results in box-plots. We observed T-LEAF improves action prediction by an average of $\approx 3.3-3.4\%$. The difference in average ingredient prediction accuracy is smaller ($\approx 1.0-1.5\%$), possibly due to the logic statements being defined over actions, e.g., the ordering temporal logic was only relevant for actions. Nevertheless, we see improvements over the reference models simply by incorporating temporal logic via our embedder. The relative differences are also consistently higher than the model checker, suggesting that learned embeddings can be more effective at training models to use prior knowledge.

\begin{figure*}
      \centering
      \includegraphics[width = 0.95\textwidth]{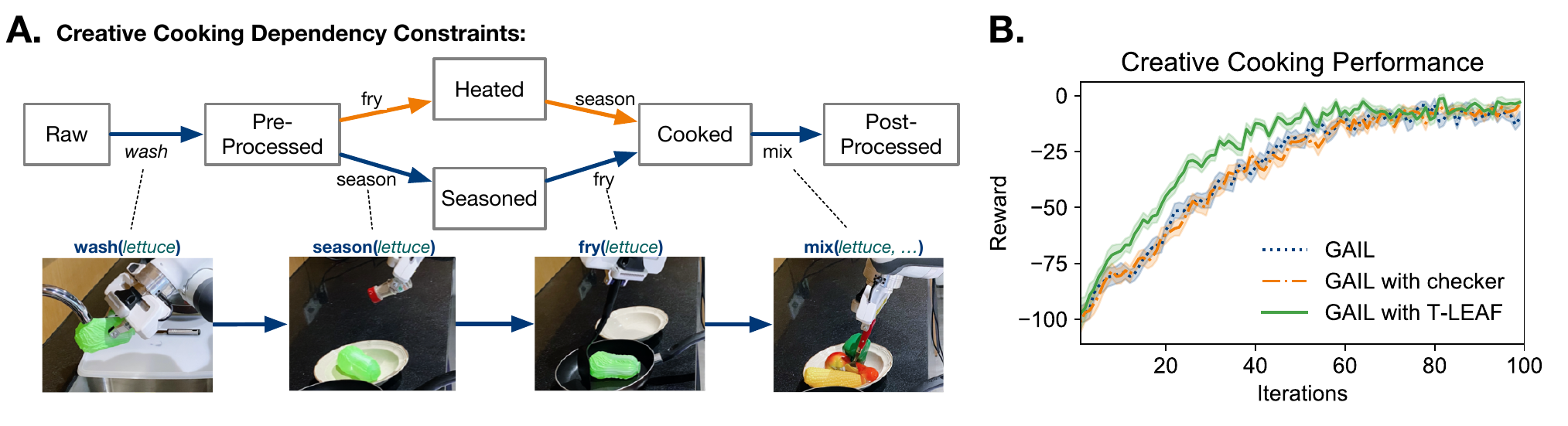}
      \caption{\small Creative Cooking Imitation Learning Experiment. (\textbf{A.}) The dependency constraints in the environment, along with a sample sequence of actions executed by the robot. The ingredient (\textit{lettuce}) changes its status from \textit{raw} to \textit{post-processed} as actions are performed. Note an alternative path is permitted by the constraints (the lettuce is fried before seasoned). (\textbf{B.}) T-LEAF improves the convergence rate of the GAIL reference model compared to a model-checker loss.}
      \label{fig:cooking}
\end{figure*}

\section{Imitation Learning for Creative Cooking}
\label{sec:experiment_imitation}

Our second experiment involves imitation learning in an object-manipulation and temporal-reasoning environment  that we call ``Creative Cooking''. The robot starts with a set of randomly sampled raw ingredients (e.g. \textit{apple} and \textit{corn}) and is tasked to prepare a meal. Imitation learning is useful in this scenario where the reward function is difficult to define; the evaluation criteria are complex and diverse. 

Unlike the previous experiment, our goal is to obtain a \emph{policy} $\pi(a_t|s_t)$, rather than a sequence model. The robot has access to the environment and a set of expert trajectories.  As in many settings, the number of expert demonstrations is small due to high collection cost. We assume that the robot cannot query the expert for more data nor directly observe the reward signals.
However, in addition to demonstrations, the experts could provide domain knowledge, high-level guidance, and preferences, which may be hard to learn from limited demonstrations. These type of knowledge could be expressed as \LTLf{}, and integrated into our policy via T-LEAF. Due to space constraints, we focus on the key concepts and relegate details to the appendix.

\mypara{Environment. }
We model Creative Cooking as a discrete Markov Decision Process (MDP) where the 
each state comprises the sampled ingredients and their corresponding status. There are 50 ingredients and 15 ingredient properties (e.g., \textit{liquid}, \textit{seasoning}). Each ingredient has at least one \textit{property} and one of six possible \textit{statuses}. 
All ingredients start from the \textit{raw} status. At each step, the robot chooses one out of 31 possible actions (e.g. \textit{wash}, \textit{cut}, \textit{cook}) and the ingredient(s) to apply the action on. 
Each action is associated with affordance constraints and pre-requisite status requirements (e.g., a carrot needs to be in a \textit{pre-processed} status before cooking). Failing the meet the affordance constraints or the pre-requisite requirements renders the action infeasible and state remains unchanged. Otherwise, the ingredient will change to a specified status. The task goal is to change all ingredients to their required status, which depends on ingredient properties. 


\mypara{Temporal Knowledge and Hierarchical Embedder. } The propositions here are action ingredient combinations; 
the proposition, \textit{action}(\textit{ingred}) at time step $t$ is \textsf{True} if it's selected by the robot at time $t$, and \textsf{False} otherwise. The expert provides affordance constraints and the dependency relationships (Fig. \ref{fig:cooking}.A.) expressed in \LTLf{}, which  comprises our knowledge base. 

Our embedder structure is similar to the previous experiment.
As the training dataset, we randomly sample 300 sub-formulae from the knowledge base along with 10 satisfying assignments and 10 unsatisfying assignments. As before, the embedder is trained and fixed before imitation learning.

\mypara{Target Model and Training via T-LEAF. }  We use GAIL~\cite{Ho2016} as the imitation learning reference method. During training, we extract the related clauses for the current ingredient to form the DFA $A_\varphi$, which is then used in the logic loss. The policy is trained using a linear combination of the logic loss and the GAIL discriminator loss with $\lambda=1.0$.%



\mypara{Results. }
Fig. ~\ref{fig:cooking}.B. summarizes our results and shows the accumulated rewards across the training iterations (averaged over 20 independent runs, with standard-errors shaded). The final performance of the models are similar, but GAIL with T-LEAF loss converges faster compared to a model checker loss. These results again support the notion that DFA embeddings can be used to train deep models. Note also that the formula related trajectories were previously \emph{unseen} by T-LEAF embedder, which suggests T-LEAF is able to generalize to unseen formulae. 

\section{Conclusion and Future Work}
\label{sec:conclusion}
This paper proposes T-LEAF, a novel approach for utilizing symbolic temporal logic by embedding \LTLf{}  DFA via graph neural networks. Empirical results are promising: T-LEAF improved deep models for sequential human action recognition and imitation learning. We believe T-LEAF is a first-step towards an alternative approach for training deep models to be compliant with pre-existing knowledge. We look forward to future improvements and applications in domains where utilizing prior knowledge can be beneficial, such as medical and human-collaborative robotics. 






\balance 
\bibliographystyle{IEEEtran}
\bibliography{ref}
\newpage

\section*{Appendix}

In the following, we provide additional details on the Creative Cooking experiment.

\mypara{MDP States.}
The environment states consist of both sampled ingredients and their corresponding status. 
There are 50 ingredients and 15 ingredient properties (e.g., \textit{liquid, seasoning}). Each ingredient has at least one property and exactly one status. Properties are used to define and determine action feasibility and environment transitions, and are not reflected in state representation. Each item's status starts from \textit{raw} and changes depending on the robot actions and environment constraints.
There are six possible statuses: \textit{raw, pre-processed, heated, seasoned, cooked, post-processed}. The transitions between these different status is illustrated in Fig. \ref{fig:cooking}.A.

At initialization of each trajectory (environment reset), 5 ingredients sampled from ingredients set uniformly without replacement, together with a special ingredient \textit{mixture} which handles ingredients merging, composes \textit{trajectory ingredients}. The \textit{trajectory ingredients}'s status are initialized to \textit{raw}, and  change via actions as described below. 
Properties of each ingredient, except \textit{mixture}, remain unchanged by actions. As such, the state representation comprises the 6 \textit{trajectory ingredients} and their respective statuses, resulting a 12-dimensional vector, i.e. \textit{(mixture idx, mixture status idx, ingred$_1$ idx, ingred$_1$ status idx, ..., ingred$_5$ idx, ingred$_5$ status idx)}.

\mypara{MDP Actions.}
At each step, the robot chooses one action and the ingredient(s) to apply the action on. 
There are 31 possible actions classified to 10 categories (e.g. \textit{cut, wash, cook}). 
All actions are applied on exactly one ingredient, except action \textit{top-with} and \textit{combine} which are applied on two ingredients. These actions (together with \textit{add}) are specially designed to handle the merging of ingredients: 
\begin{itemize}
    \item \textit{add(ingred)}: Add \textit{ingred} in \textit{mixture}. The properties of \textit{ingred} will be appended to \textit{mixture}. The original position of \textit{ingred} in the state representation will be replaced by zeros, which represents that the ingredient no longer exists separately from the mixture.
    \item \textit{combine(ingred$_1$, ingred$_2$)}: The effect is equivalent to \textit{add(ingred$_1$)} and \textit{add(ingred$_2$)}, i.e. add \textit{ingred$_1$} to \textit{mixture} then add \textit{ingred$_2$} to \textit{mixture}.
    \item \textit{top-with(ingred$_1$, ingred$_2$)}: Top \textit{ingred$_1$} with \textit{ingred$_2$}. \textit{ingred$_2$}'s position in the state representation will be replaced by zeros, while the status of \textit{ingred$_1$} will change appropriately (its properties remain unchanged). For example, topping meat with salt changes the status of the meat to \emph{seasoned}. 
\end{itemize}

\mypara{MDP Transitions}
Each action is associated with affordance constraints and pre-requisite status requirements. Failing the meet either of these renders the action infeasible and state remains unchanged. Otherwise, the ingredient will change to a specified status.

Affordance constraints are common-sense facts about whether actions can only or cannot be applied to some ingredients (e.g., we cannot cut milk). 
Pre-requisite status requirements refer to status dependencies (Fig. \ref{fig:cooking}.A.) (e.g., a carrot needs to be in a \textit{pre-processed} status before cooking).
Our temporal knowledge consists of both affordance constraints and pre-requisite status requirements, i.e., the dependency relationships.

\mypara{MDP Rewards}
The task goal is to change all ingredient to their required status, which depends on ingredient properties. 
The robot gets a large positive reward, $+20$, if the goal is reached. There is a $-1$ penalty for each step taken and an extra $-1$ penalty if the action chosen is infeasible.

\end{document}